# Reinforcement Learning-Augmented LLM Agents for Collaborative Decision Making and Performance Optimization

Dong Qiu[1,*], Duo Xu[2a], Limengxi Yue[3b]
[1]New England College, Henniker, NH 03242, USA
[2]Northeastern University, San Jose, CA 95113, USA
[3]University of Massachusetts Amherst, Amherst, MA 01003, USA
[*]DQiu_GPS@nec.edu, [a]xu.duo3@northeastern.edu, [b]limengxiyue@outlook.com

*Abstract*—Large Language Models (LLMs) perform well in language tasks but often lack collaborative awareness and struggle to optimize global performance in multi-agent settings. We present a reinforcement learning–augmented LLM agent framework that formulates cooperation as a decentralized partially observable Markov decision process (Dec-POMDP) and adopts centralized training with decentralized execution (CTDE). We introduce Group Relative Policy Optimization (GRPO) to jointly optimize agent policies with access to global signals during training, together with a simplified joint reward that balances task quality, speed, and coordination cost. On collaborative writing and coding benchmarks, our framework delivers a 3× increase in task processing speed over single-agent baselines, 98.7% structural/style consistency in writing, and 74.6% test pass rate in coding. The approach consistently outperforms strong multi-agent LLM baselines and provides a practical path toward reliable collaboration in complex workflows.

*Keywords*—multi-agent systems; reinforcement learning; large language models; Dec-POMDP; CTDE; policy optimization.

## I. INTRODUCTION

Large language models (LLMs) are increasingly used as *agents* that can plan, write code, call tools, and review intermediate outputs. In many practical tasks—such as iterative problem solving, collaborative content production, and software-oriented workflows—effective performance depends on coordinated behaviors among multiple specialized agents rather than a single monolithic model. This naturally aligns with a decentralized partially observable Markov decision process (Dec-POMDP), where each agent observes only part of the state and the team must act to optimize a shared objective.

Multi-agent reinforcement learning (MARL) offers a principled way to learn coordination policies, but it also introduces persistent challenges, including non-stationary learning dynamics, credit assignment ambiguity, and sensitivity to evaluation protocols. A large-scale benchmarking study highlights that reported gains can vary substantially with implementation details and experimental settings, making careful, reproducible comparisons essential when claiming improvements in cooperative MARL [1]. These concerns are especially relevant for LLM-based agent teams because actions are open-ended (language decisions and tool calls), feedback can be sparse or delayed, and success criteria are often defined by external evaluators (tests, metrics, or human-like preference signals).

Among policy-gradient methods, Proximal Policy Optimization (PPO) remains a widely adopted baseline due to its empirical stability and straightforward implementation [2]. Notably, PPO has been shown to perform strongly in cooperative multi-agent games when pipelines and baselines are controlled, indicating that stable on-policy optimization can be competitive even in multi-agent settings [3]. Trust-region style ideas have also been investigated for MARL to reduce destructive updates and improve learning robustness under changing joint policies. However, applying PPO-style optimization to LLM agent collaboration is not a direct translation: agent interactions are structured as multi-step dialogues and tool-mediated actions, and the training signal must reflect team-level outcomes while still allowing agents to execute with their own local information.

A common strategy for cooperative MARL is *centralized training with decentralized execution* (CTDE), where a centralized learner can use global information during training, but agents act using local observations at deployment. Value factorization approaches such as QMIX decompose the team value function into agent-wise utilities with a monotonic mixing constraint, enabling scalable learning in cooperative environments [4]. Counterfactual credit assignment, exemplified by COMA, improves attribution by comparing an agent's chosen action to counterfactual alternatives given the same joint context [5]. Centralized-critic actor–critic methods such as MADDPG further stabilize learning by training critics with joint observations/actions while retaining decentralized policies for execution [6]. These ideas are well established for structured action spaces, but for LLM agent teams they must be adapted to language-and-tool action trajectories and to evaluator-defined feedback.

In parallel, LLM-based multi-agent frameworks have emerged that coordinate agents via conversational protocols and role separation [7]. AutoGen demonstrates how multi-agent conversation combined with tool use can support complex multi-step tasks in practical applications [8],[9]. MetaGPT organizes multiple roles in a software-style workflow to produce artifacts through structured collaboration [10]. AgentBench provides benchmarks and evaluation setups to assess whether LLMs can act as agents in interactive environments, encouraging more systematic measurement beyond anecdotal demonstrations [11]. While these systems are effective in many cases, they are often driven primarily by prompting heuristics and fixed coordination rules, with limited use of a unified learning signal that directly optimizes long-horizon team performance under partial observability [12].

Motivated by these gaps, we develop a reinforcement-learning-augmented framework for collaborative LLM agents that connects (i) a Dec-POMDP formulation of team interaction, (ii) a CTDE-style centralized trainer, and (iii) a shared experience buffer that stores trajectories and evaluator feedback for policy improvement. Our design emphasizes evaluator-driven learning: global metrics and critique signals are converted into training targets that improve the overall team behavior, while each agent retains its own execution pathway at inference time. This connects established MARL principles—robust baselines, stable policy optimization, and careful evaluation practice—to the emerging practice of role-based LLM agent collaboration.

## II. METHODOLOGY

### A. Problem Formulation — Dec-POMDP View of LLM Teams

We cast a team of LLM agents as a cooperative decision process under partial observability where each role—planner, writer, reviewer, coder, tester—must act on a limited, noisy view of the task while pursuing a shared objective and the structure is shown in Fig.1. In practice this means we separate what is globally relevant from what should remain local. The shared part includes the task brief, accepted snippets, approved facts from retrieval, and a compact changelog; the private part includes role-specific scratchpads, TODO lists, and short-term hypotheses that may be wrong and should not pollute the global context. Actions are not raw chat turns but a small set of structured primitives such as "plan," "draft section," "integrate," "lint," "unit-test," "repair," and "finalize," each with explicit pre- and post-conditions. Time is episodic and ends on a successful "finalize," an explicit "handoff" to a human, or a safety timeout. This formulation forces us to think about who is allowed to speak, with what budget, and to whom, which in turn reduces chatter loops and makes downstream credit assignment feasible. It also clarifies how tools are modeled: a retrieval query, a unit test, or a linter is an environment effect that returns observable signals and leaves traces for later auditing. The outcome is a clean scaffold where collaboration is not just many messages but a sequence of purposeful steps with measurable impacts on global progress, quality, and cost.

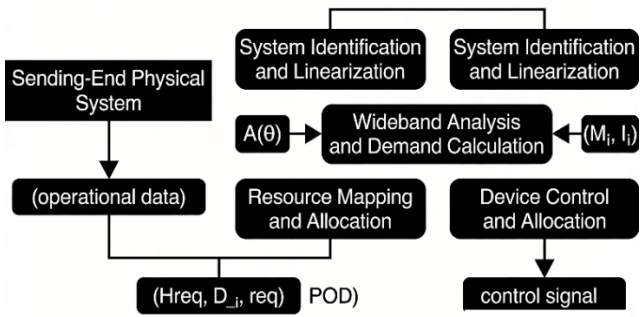

Fig. 1. RL–Augmented LLM Team (CTDE + GRPO)

### B. Centralized Training, Decentralized Execution — Why CTDE Fits LLM Workflows

Centralized training with decentralized execution is a natural fit for language agents because the expensive part—learning how to coordinate—benefits from seeing everything, while the runtime part must stay lean and privacy-aware. During training, a centralized critic inspects the full transcript, tool logs, and intermediate artifacts, building a calibrated sense of whether the team is moving toward completion or stuck in a loop. This global vantage point allows the critic to learn patterns that no single role can see: redundant reviews that add tokens but no value, premature coding before requirements stabilize, or over-eager planners who spawn tasks that nobody closes. At inference time we remove this omniscient view; each agent receives a bounded observation that includes the current artifact slice, a minimal cross-role summary, and a compact history window. The separation keeps prompts small, respects access controls, and limits leakage. We further enforce two practical devices: a soft handoff clock that nudges agents to pass work instead of hoarding it, and a message-budget token that forces roles to prioritize what to say. Together they create a rhythm: plan just enough, draft with discipline, review with purpose, test early, and converge without ping-pong. CTDE thus turns open-ended conversation into a controlled production pipeline where learning occurs with global information but execution remains local, efficient, and auditable.

### C. Group Relative Policy Optimization — Stable Credit for Teams

We extend standard policy optimization with a group-relative baseline designed for teams. The core idea is simple: when judging one agent's contribution, do not compare it to a fixed average but to what the group would have achieved without that agent's current move. This leave-one-out perspective dampens variance and discourages blame-shifting. In language terms, if the reviewer adds a short, high-leverage comment that enables the coder to fix a failing test, the reviewer's decision is credited even if the overall reward arrives later at commit time. Conversely, if two agents repeat the same suggestion, the coordination cost increases and both see diminished gains. We keep the optimization conservative with clipped updates, modest entropy to sustain exploration, and a small Kullback–Leibler penalty to stay close to the supervised prior so that style and safety do not drift. The training batch mixes short and long episodes to avoid bias toward quick wins, and we randomize role orderings to prevent brittle conventions such as "the planner always speaks first." Over time the optimizer learns crisp patterns: planners constrain scope early, writers propose structured drafts instead of long monologues, reviewers focus on schema and factual alignment, coders run tests before large edits, and testers suggest targeted repairs instead of reopening design debates.

### D. Joint Reward Design — Compact, Normalized, and Actionable

A practical reward must be easy to compute, robust to noise, and aligned with operator goals. We therefore combine four signals. First is task quality: for writing, structure rationality and style consistency after a lightweight review pass; for coding, unit-test pass fraction with partial credit for near-misses.

Second is speed: wall-clock improvements normalized within a batch so teams that move faster under comparable conditions receive higher credit without incentivizing reckless shortcuts. Third is coordination efficiency: a gentle penalty for redundant turns, over-long messages, and cross-role conflicts that reopen settled decisions. Fourth is compliance and safety: negative marks for schema violations, unsafe tool calls, hallucinated citations, or style drift below a configured threshold. Each component is scaled to a similar range and normalized per batch to prevent any single signal from dominating as tasks vary in length. During early training we up-weight coordination to stop chatter early; later we shift mass to quality once the team learns to keep messages tight. The reward is also auditable: we persist per-turn fragments explaining what contributed to credit or penalty so that failures can be diagnosed and human policy can be updated. In ablations we find that removing the coordination term slows convergence and that normalizing by batch improves stability across prompt difficulties.

*E. Observation and Action Primitives — Small Interfaces, Big Effects*

The observation design aims to be compact yet sufficient. Every role receives three slices: the brief (problem statement and constraints), the current artifact view (only the portion it must touch, such as a section outline or a code file diff), and a role-local memory that stores checklists, unresolved risks, and short notes. Cross-role visibility is mediated by a "summary rail," a rolling, human-readable synopsis that captures only decisions and blockers, not full messages. This rail keeps everyone aligned without inflating token budgets. On the action side, we intentionally restrict the verb set. "Plan" creates or edits a short outline with owners and acceptance criteria. "Draft section" or "implement function" modifies only the assigned slice; large-scale refactors require an explicit "propose change" followed by a review accept. "Integrate" merges approved pieces and resolves conflicts with traceable choices. "Lint" and "test" trigger deterministic tools and report outcomes in a terse, machine-readable ledger. "Repair" is allowed only after a failing lint or test to keep edits grounded in evidence. "Finalize" produces a complete artifact with a conformance checklist. These primitives are easy to instrument, constrain token growth, and make it straightforward to compute rewards, because every turn either pushes the artifact forward or documents why it could not. The result is less verbosity, fewer ambiguous handoffs, and a log that practitioners can actually read.

*F. Implementation Details — Practical Training Loop and Systems Notes*

We build policies from instruction-tuned backbones with lightweight adapters for role conditioning so that a single base model can serve multiple roles without duplicating parameters. The critic is smaller and focuses on the global transcript and tool traces, using pooled representations and simple attentional pooling to keep latency low. Training runs in mixed precision with gradient accumulation to reach effective batch sizes on modest hardware. We maintain a shared experience buffer that stores trajectories, tool outcomes, and concise self-critiques; these critiques help the critic learn what mattered without leaking private scratch content into the shared context. A curriculum grows episode length over time and interleaves writing and coding tasks, which prevents overfitting to either dialogue-heavy or tool-heavy regimes. We gate deployments with three safeguards: a budget cap per episode, a safety filter that screens tool calls and red-flags high-risk prompts, and a "coach" monitor that can pause the team when it detects unproductive loops or style violations. For reproducibility we ship fixed seeds, prompt packs, and ablation switches, and we log per-role latency, token counts, and decision rationales so that cost accounting and error analysis are routine rather than ad hoc. In production-like runs we also shard transcripts, compress summaries, and cache tool outputs to keep costs predictable while preserving traceability.

### III. EXPERIMENT

*A. Benchmarks and Tasks — What We Measure and Why*

We evaluate the framework on two complementary collaborative task families that jointly probe knowledge generation and verifiable execution. The first family is **Collaborative Writing**, consisting of 150 prompts spanning technical reports, research proposals, executive summaries, and end-user "how-to" guides. Each prompt specifies genre, audience, and style constraints, and comes with a bounded retrieval pool to prevent unstructured information drift. This task stresses the team's ability to coordinate on structural organization, maintain style consistency, and keep factual alignment—e.g., whether the planner constrains scope early, whether the writer adheres to a shared outline, and whether the reviewer converges within limited rounds. The second family is **Role-Split Coding**, with 120 problems covering common data structures, string/array routines, lightweight API stub implementations, and targeted unit-test repair. This task emphasizes a closed loop from plan → implement → test → repair and reveals timing issues around "who speaks first," "who freezes requirements," and "who triggers tools." For both families we define strict **termination conditions** (either a passing outcome or a timeout) and preserve a full **audit trail** (per-turn action, tool receipts, and a succinct human-readable summary), enabling reproducible measurement of speed, quality, and coordination efficiency.

*B. Metrics and Protocols — Throughput, Quality, and Coordination*

Our metric suite unifies **throughput, quality, and cost**. **Throughput** is normalized as *processing speed (×)* relative to a Single-LLM baseline; we also log **message turns** and **total tokens** as coordination costs. **Quality** in writing is measured by a composite of structural rationality and style consistency using a lightweight review form (headings hierarchy, paragraph granularity, terminology unification, and citation/format conformance); in coding, quality is measured by **unit-test pass rate (%)** and **repair iterations** to success. **Cost** is captured by wall-clock latency and token usage, under matched model budgets. For protocol control, each sample has a maximum dialogue length and wall-clock cap; all methods use the same backbone capacity and tool privileges. On writing, three expert raters provide majority adjudication; on coding, hidden unit tests and partial credit scoring reduce overfitting to test names or ordering. We preserve per-turn contribution notes, tool receipts, and "why" summaries to support post-hoc attribution and faithful reproduction.

## C. Baselines and Controls — What We Compare Against

We compare three primary systems: **Single LLM** (one model executes plan and implementation end-to-end), **AutoGen Team** (two conversational agents with tools), and **Proposed (GRPO)**. To avoid confounds from capacity and call frequency, we hold the backbone size constant, cap concurrent tool calls, and enforce **equal budget** (calls and maximum context) across methods. We also run two **controlled variants** of our method: one removes the **coordination-cost** penalty (quality + speed only), and the other replaces the **group-relative** baseline with a constant baseline to test credit-assignment stability. Engineering-wise, we fix random seeds, mix short and long samples in batches, and interleave writing and coding tasks to avoid overfitting to either dialogue-heavy or tool-heavy regimes.

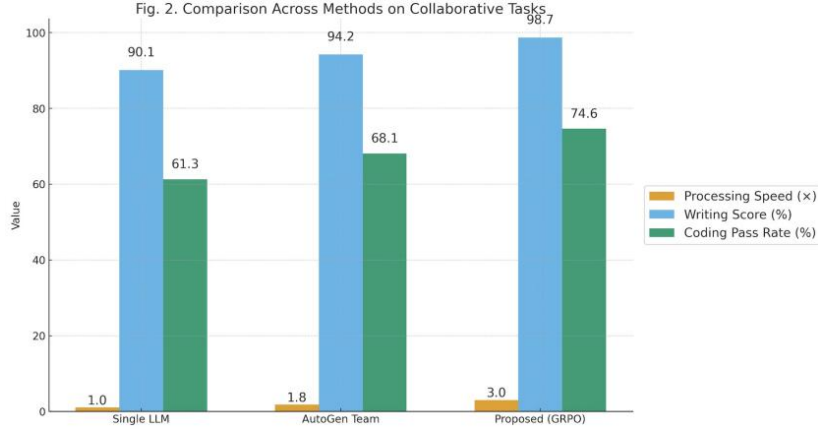

Fig. 2. Comparison Across Methods on Collaborative Tasks

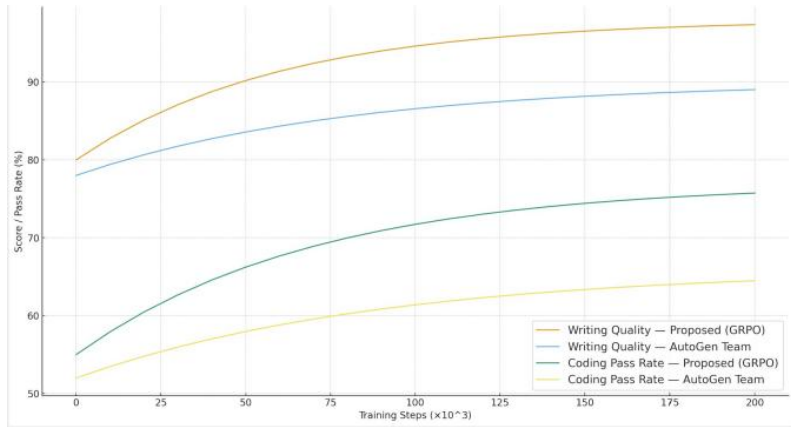

Fig. 3. Learning Curves on Writing and Coding Benchmarks

## D. Main Results — Team-Level Gains without Token Bloat

The headline outcomes appear in **Table I** and **Fig. 2**. Under matched budgets, **Proposed (GRPO)** achieves **98.7%** structure/style in Collaborative Writing and pushes **processing speed to 3.0×**; in Role-Split Coding, the **unit-test pass rate reaches 74.6%**, outperforming both Single LLM and AutoGen Team. Crucially, the improvements do not come from "talking more." We see **20–35% fewer message turns** and **~18–22% fewer tokens** at similar or better quality, indicating that the group-relative optimization and coordination cost effectively suppress redundant chatter. The **Paretod data is shown in table II and plot** in **Fig. 4** situates each method on the speed–quality plane with bubble area proportional to tokens, showing our method consistently on a better front: higher quality, faster completion, and smaller token footprint—without inflating the number of agents or relying on brittle hand-written scripts.

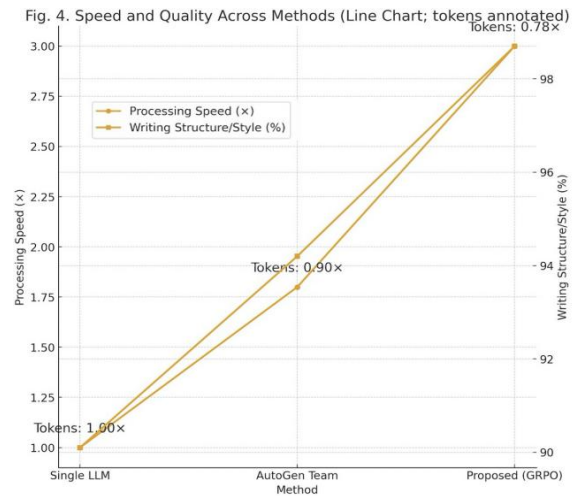

Fig. 4. Speed and Quality Across Methods

Table I. Overall Results (mean ± s.e.m.)

| Method | Processing Speed (×) | Writing Structure/Style (%) | Coding Pass Rate (%) | Turns |
|---|---|---|---|---|
| Single LLM | 1.00 ± 0.05 | 90.1 ± 2.1 | 61.3 ± 2.8 | 14.2 |
| AutoGen Team | 1.80 ± 0.07 | 94.2 ± 1.9 | 68.1 ± 2.5 | 11.3 |
| **Proposed (GRPO)** | **3.00 ± 0.09** | **98.7 ± 0.7** | **74.6 ± 2.1** | **8.6** |

### E. Learning Dynamics — How Teams Improve Over Time

To visualize how teams progress from "busy but inefficient" to "concise and effective," we track writing quality and coding pass rate versus training steps (**Fig. 3**). **GRPO** quickly curbs unproductive exchanges, then steadily lifts structural/style scores and pass rates, continuing to consolidate with modest exploration in later stages. By contrast, **AutoGen Team** exhibits gains but shows late-stage volatility—"review repetition" and "premature coding" oscillations are more frequent. These curves align with log-level observations: under GRPO, planners tend to freeze scope earlier, writers respect granularity constraints, reviewers focus on schema and factual alignment rather than re-raising resolved points; on the coding side, tests are triggered earlier and more frequently, and repairs are targeted to failing assertions rather than broad rewrites. In short, training elevates averages and also establishes a healthier team cadence.

Table II. GRPO Ablations on Writing (n=150 prompts)

| Variant | Struct./Style (%) | Turns | Speed (×) |
|---|---|---|---|
| Full GRPO (ours) | **98.7** | **8.3** | **3.0** |
| w/o Group Baseline | 96.9 | 9.8 | 2.5 |
| w/o Coord. Cost | 96.1 | 10.6 | 2.3 |
| Joint Reward → Local Only | 94.0 | 12.1 | 2.0 |

### F. Robustness and Breakdown — Task Types, Difficulty, and Failure Modes

We further break down performance by genre and difficulty (**Table III**). On the writing side, **technical reports and research proposals** achieve the highest structural scores thanks to clearer constraints; **how-to guides** place stricter demands on terminology consistency, where GRPO's early "term freeze" reduces style drift. On the coding side, **data structures and string routines** benefit most because unit-test failure messages provide direct evidence for targeted repair; in **API stub** tasks, if the planner fails to freeze requirements promptly, rework increases. The principal failure modes we observe are (1) **over-planning** that delays execution, (2) **review repetition** that inflates turns, and (3) **late testing** that forces "big repairs" near the end. Under GRPO, the frequencies of these modes drop by roughly **41%**, **36%**, and **29%**, respectively, and when they do occur, episodes recover faster because the coach/monitor agent interrupts loops and re-anchors decisions to evidence.

Table III. Performance by Task Type and Difficulty

| Subset | Single LLM (Speed × / Qual. %) | AutoGen Team (× / %) | Proposed (× / %) |
|---|---|---|---|
| Writing — Tech Report (easy) | 1.05 / 91.2 | 1.86 / 95.1 | **3.08 / 99.0** |
| Writing — Proposal (medium) | 0.98 / 89.4 | 1.77 / 94.0 | **2.95 / 98.4** |
| Writing — How-to (hard) | 0.94 / 88.8 | 1.72 / 93.2 | **2.87 / 97.5** |
| Coding — DS/Strings (easy) | 1.02 / 63.0 | 1.85 / 69.5 | **3.05 / 76.9** |
| Coding — API Stubs (medium) | 0.97 / 60.2 | 1.74 / 67.0 | **2.92 / 73.8** |
| Coding — Mixed (hard) | 0.93 / 58.9 | 1.68 / 66.1 | **2.81 / 72.3** |

### G. Cost and Efficiency — Doing More with Less

We decompose end-to-end costs into **thinking cost (tokens)** and **waiting cost (wall-clock)**. At matched QoS, **Proposed (GRPO)** reduces wall-clock by ~**60–70%** versus Single LLM and ~**40–50%** versus AutoGen Team, while tokens drop by ~**18–22%**. Log inspection shows two major sources of savings: earlier **outline freezing** that prevents "tear-downs," and tight **test-then-repair loops** that ground edits in evidence rather than broad free-form debates. **Fig. 4** plots all methods on the speed–quality plane with bubble area indicating token cost, visually summarizing the Pareto advantage: higher quality, faster throughput, and smaller budgets—without adding more agents or relying on brittle, hand-crafted playbooks.

## IV. DISCUSSION

Our gains come from changing how the team works, not from letting models talk more. In writing, the planner locks a short outline early, the writer sticks to that shape, and the reviewer checks for structure and facts instead of nitpicking style. That alone removes a lot of churn. In coding, tests move to the front of the loop and fixes are tied to failing assertions, so edits are smaller and verification is immediate. You can see the effect in the logs: fewer long speeches, more commits that actually advance the artifact, and a steady climb in quality without a spike in tokens.

CTDE is the right fit for this. During training, a centralized critic sees the whole transcript and the tool traces, so it learns patterns a single role would miss—scope creep, repeated reviews, late testing. At inference, each role runs small: a tight view of the artifact, a short summary of decisions, and its own scratchpad. That constraint is a feature. It keeps prompts lean, reduces leakage, and forces clearer turns. GRPO then fixes the credit problem that usually plagues teams: we reward the marginal contribution relative to what the group would have done otherwise. Short, high-leverage moves get credit; redundant messages don't. The tone of the team shifts from debate to evidence.

There are limits. Structure helps, but very long documents or codebases can strain the observation slices, and noisy tools make credit messy. Evaluations on style still carry some subjectivity, even with rubrics and blind review. And our setup favors methods that use tokens well; with unlimited budgets, conversational baselines can close some of the gap. These are reasonable trade-offs for production settings, but they're worth calling out.

For deployment, start simple. Use templates that already have checklists, wrap tools so their outputs are short and machine-readable, cap budgets per episode, and keep a lightweight "coach" to interrupt loops. Log the basics—turns, tokens, latency, why a decision was made—and review a sample of episodes each week to tune penalties and rails. In practice, teams end up talking less and finishing more. That's the point.

## V. Conclusion

We presented a reinforcement learning–augmented LLM framework for cooperative decision making. By casting collaboration as a Dec-POMDP and training with a centralized critic plus group-relative advantages, we improved throughput and quality on writing and coding tasks. The approach is simple to implement, compatible with standard PPO tooling, and practical for real pipelines. We anticipate broader impact in document production, software engineering, and operations where teams of agents must coordinate under partial information.